\documentclass[12pt,aps,showpacs]{revtex4}%
\usepackage{amsmath}
\usepackage{amsfonts}
\usepackage{amssymb}
\usepackage{graphicx}%
\providecommand{\U}[1]{\protect\rule{.1in}{.1in}}
\begin{document}
\title{Model-Driven Applications of Fractional Derivatives and Integrals}
\author{William A. Sethares}
\affiliation{Department of Electrical and Computer Engineering, University of Wisconsin,
Madison, USA}
\email{sethares@ece.wisc.edu}
\author{Sel\c{c}uk \c{S}. Bay{\i}n }
\affiliation{Institute of Applied Mathematics, Middle East Technical University, Ankara,
Turkey }
\email{bayin@metu.edu.tr}
\date{March 17, 2014}

\begin{abstract}
Fractional order derivatives and integrals (differintegrals) are viewed from a
frequency-domain perspective using the formalism of Riesz, providing a
computational tool as well as a way to interpret the operations in the
frequency domain. Differintegrals provide a logical extension of current
techniques, generalizing the notion of integral and differential operators and
acting as kind of frequency-domain filtering that has many of the advantages
of a nonlocal linear operator. Several important properties of differintegrals
are presented, and sample applications are given to one- and two-dimensional
signals. Computer code to carry out the computations is made available on the
author's website.

\end{abstract}

\maketitle

\section{Introduction}

Fractional calculus is a convenient way of introducing memory and nonlocal
effects into models of physical systems where integer-valued derivatives and
integrals (for example, the first derivative, the double integral) are
replaced by fractionally-valued derivatives and integrals (the half-derivative
or the 1.5th integral). One advantage is that these extend local nonlinear
models to incorporate nonlocal effects \cite{OS74, RH2000}. As far back as
1695, L'H\^{o}pital asked Leibnitz \cite{das} what happens if the order $n$ is
allowed to assume fractional values such as $\frac{1}{2}$, and this began a
longstanding quest to define and make use of derivatives and integrals with
fractional orders. Over the years, a number of different definitions have been
proposed by Riemann, Caputo, Liouville, Weyl, Riesz, Feller, and Gr\"{u}nwald,
\cite{herrmann}. Fractional generalizations of some of the basic differential
equations of physics have led to new understandings of the dynamics underlying
macroscopic phenomena in a wide range of areas including anomalous diffusion
\cite{IP99, SKB2002, KST06} and quantum mechanics \cite{bayin2, bayin3,
NL2000}.

Many modern applications are data driven, and so there is a need to be able to
calculate differintegrals numerically; we call this the ``signal processing
approach.'' Some of the most common and useful signal and image processing
techniques involve linear operations such as derivatives $\frac{df}{dt}$,
$\frac{d^{2}f}{dt^{2}}$, $...$, $\frac{d^{n}f}{dt^{n}}$ and integrals
$F=\int\!f(t)dt$, $F^{2}$, $...$, $F^{n}$ and there have been recent efforts
to apply fractional generalizations in a variety of fields such as system
identification \cite{hartley}, control \cite{maione}, simulations
\cite{krishna}, and image processing \cite{mathieu, ortigueira}. A number of
these are reviewed in the next paragraph.

Using the Gr\"{u}nwald-Letnikov definition of fractional derivatives and the
Mach effect, Tseng and Lee \cite{TsengLee} introduced an image sharpening
algorithm and demonstrated the effectiveness of their method. Khanna and
Chandrasekaran \cite{KhannaChandra}, using the Gr\"{u}nwald-Letnikov
definition proposed a multi-dimensional mask that enhances the image in
several directions in one pass. Ye et. al. \cite{YeEtAl} concentrated on
identifying the blur parameters of motion blurred images and obtained better
results with fractional derivatives as opposed to the methods based on
integer-order derivatives. They showed that fractional derivatives offer
better immunity to noise and can improve ability to determine motion-blurred
direction and extent. The authors of \cite{BaiFeng, CuestaEtAl, JanevEtAl}
discussed partial differential equations and diffusion-based image processing
techniques for filtering, denoising, and restoration via fractional calculus.
Jun and Zhihui \cite{JunZhihui} discussed a class of multi-scale variational
models for image denoising. They showed that such models can improve peak
signal to noise ratio of the image and also help preserve textures and help
eliminate the staircase effect. Using Gr\"{u}nwald-Letnikov and the R-L
definitions, Pu et. al. \cite{PuEtAl} gave six differential masks towards
texture enhancement. Using fractional differentiation and integration, Yang
et. al. \cite{YangEtAl} discussed edge detection and obtained promising
results compared to conventional methods based on integer-order differential
operators in regard to detection accuracy and noise immunity. In
\cite{NakibEtAl1, NakibEtAl2}, authors discussed image tresholding based on
fractional differentiators. In regard to document image analysis,
\cite{GhamisiEtAl} introduced two new methods of image segmentation via
fractional calculus with better performances over conventianl methods. Signal
detection in fractional Gaussian noise is considered in \cite{BartonPoor} and
Prasad et. al. \cite{PrasadEtAl} gave a new method for color image encoding
using fractional Fourier transformation.

Several factors may have limited the adoption of such fractional operators
(also called \emph{differintegrals}): the profusion of (somewhat) incompatible
definitions, the difficulty of carrying out the required fractional-order
filter designs, and the lack of a clear intuitive framework. This paper
addresses these issues by transforming into the frequency domain where Riesz's
definition can be applied directly. Instead of attempting to derive the time
(or spatial) analogs of the filters, the calculations can be carried out
directly in the frequency domain. We make the argument that this approach
provides both a conceptual simplification and that it typically leads to lower
computational complexity.

\begin{figure}[ptbh]
\begin{centering}
\includegraphics[width=1.7in]{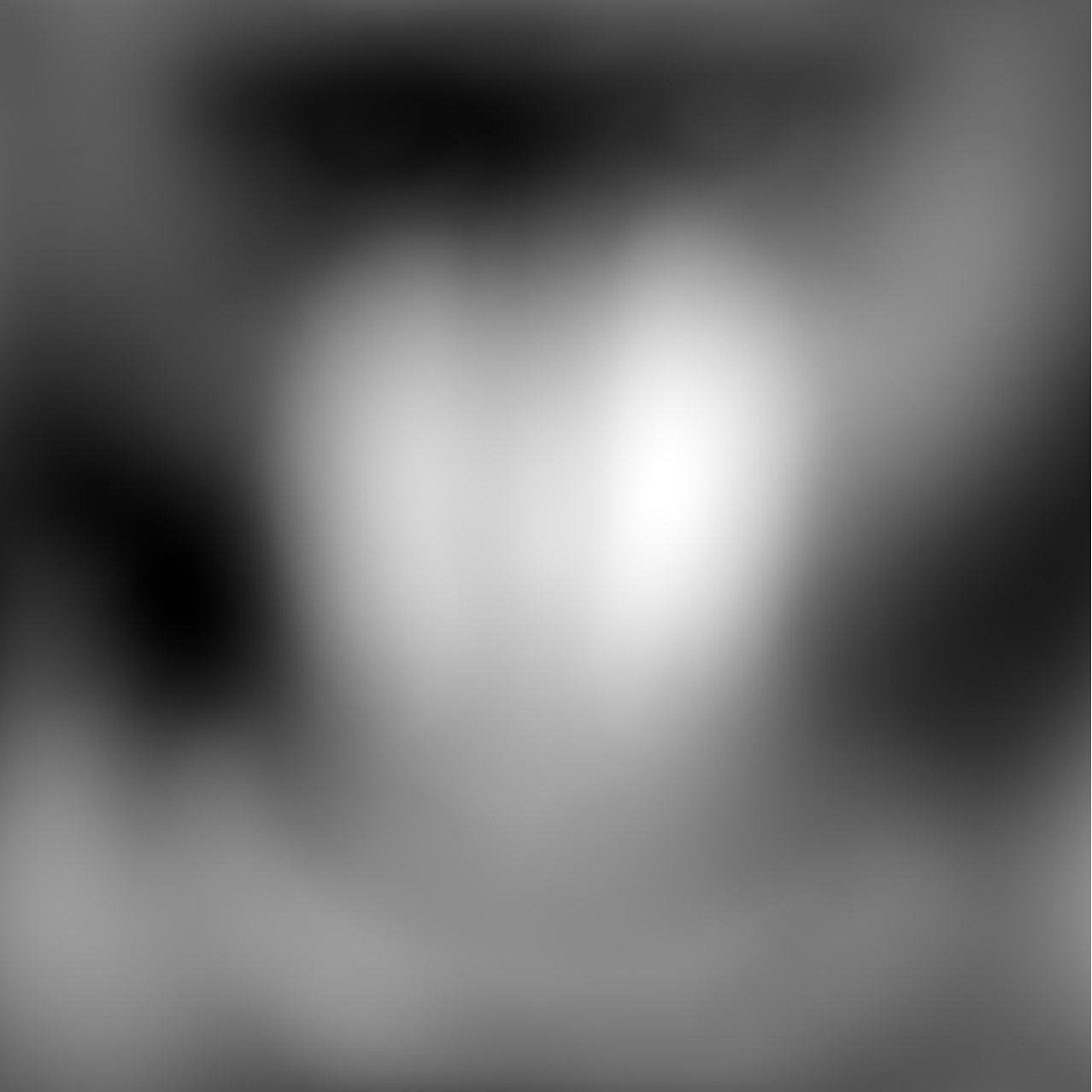} %
\includegraphics[width=1.7in]{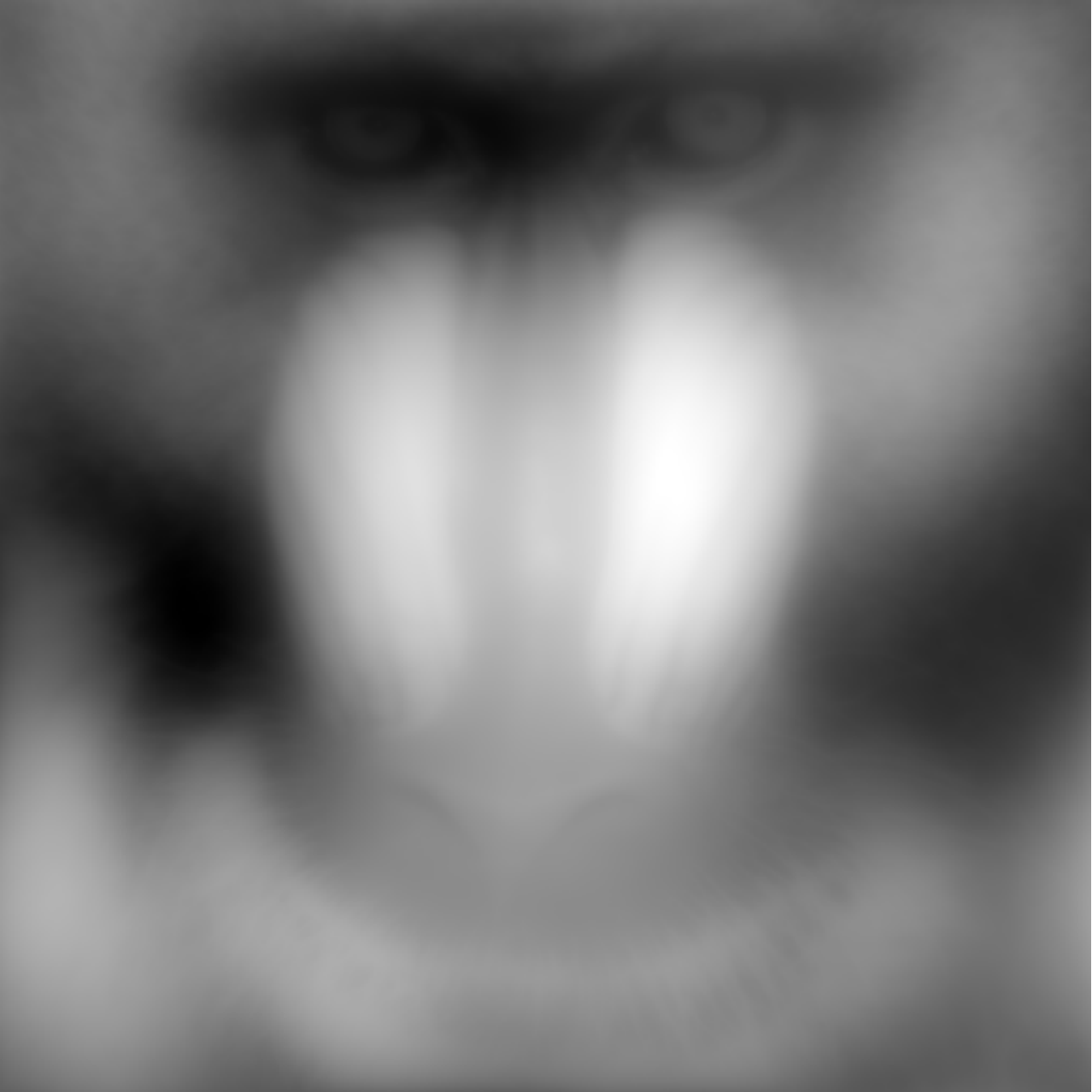}\\
\includegraphics[width=1.7in]{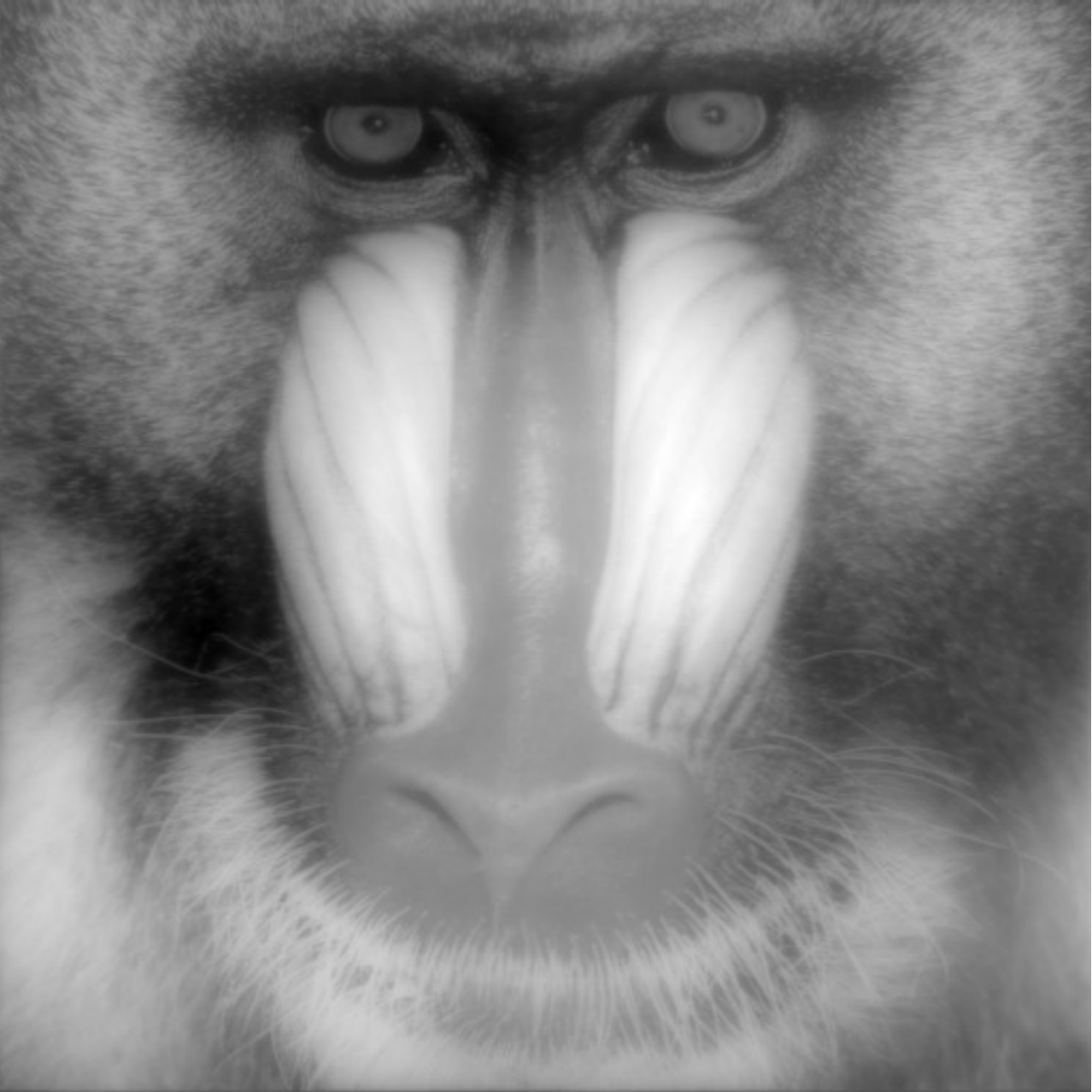}
\includegraphics[width=1.7in]{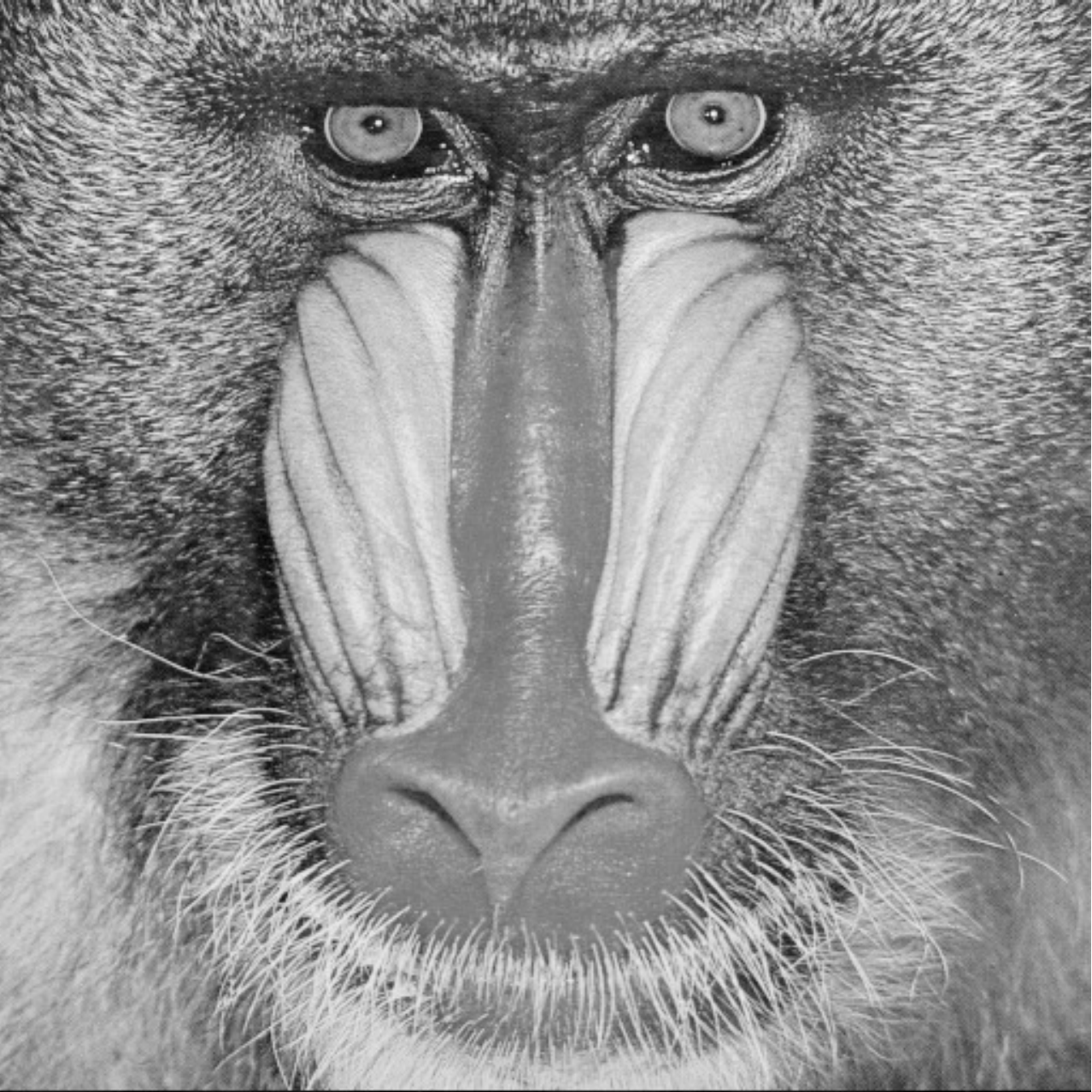}\\
\end{centering}
\caption{Fractional integration applied to the Mandrill image \cite{wolfram}
appears to blur the image somewhat analogous to the blurring of integer-valued
integrations. This shows $q=-1.8$ (upper-left), $q=-1.3$ (upper right),
$q=-0.6$ (lower left), and $q=0$ (the original image on the lower right). }%
\label{fig:mandrilInt}%
\end{figure}

\begin{figure}[ptbh]
\begin{centering}
\includegraphics[width=1.7in]{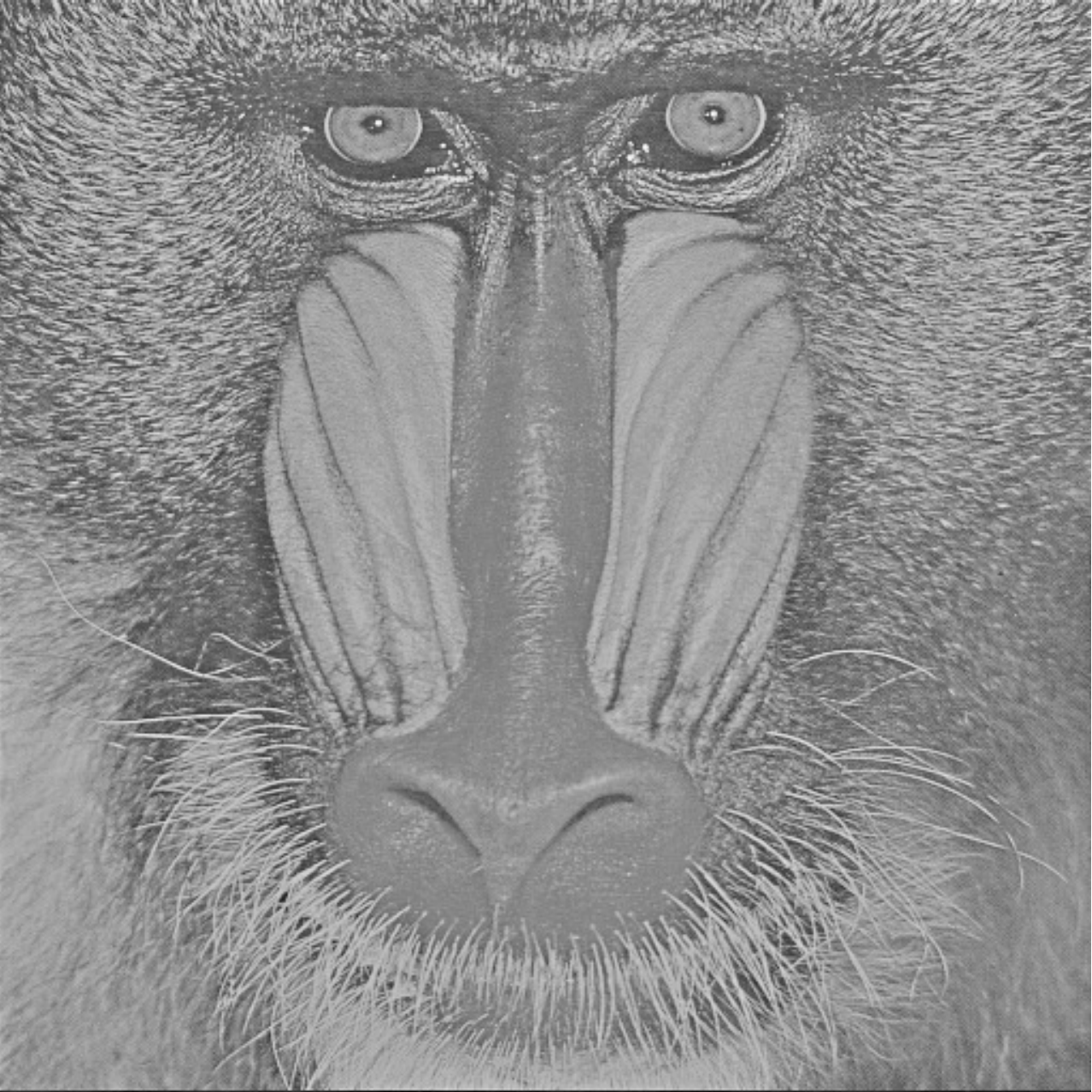}
\includegraphics[width=1.7in]{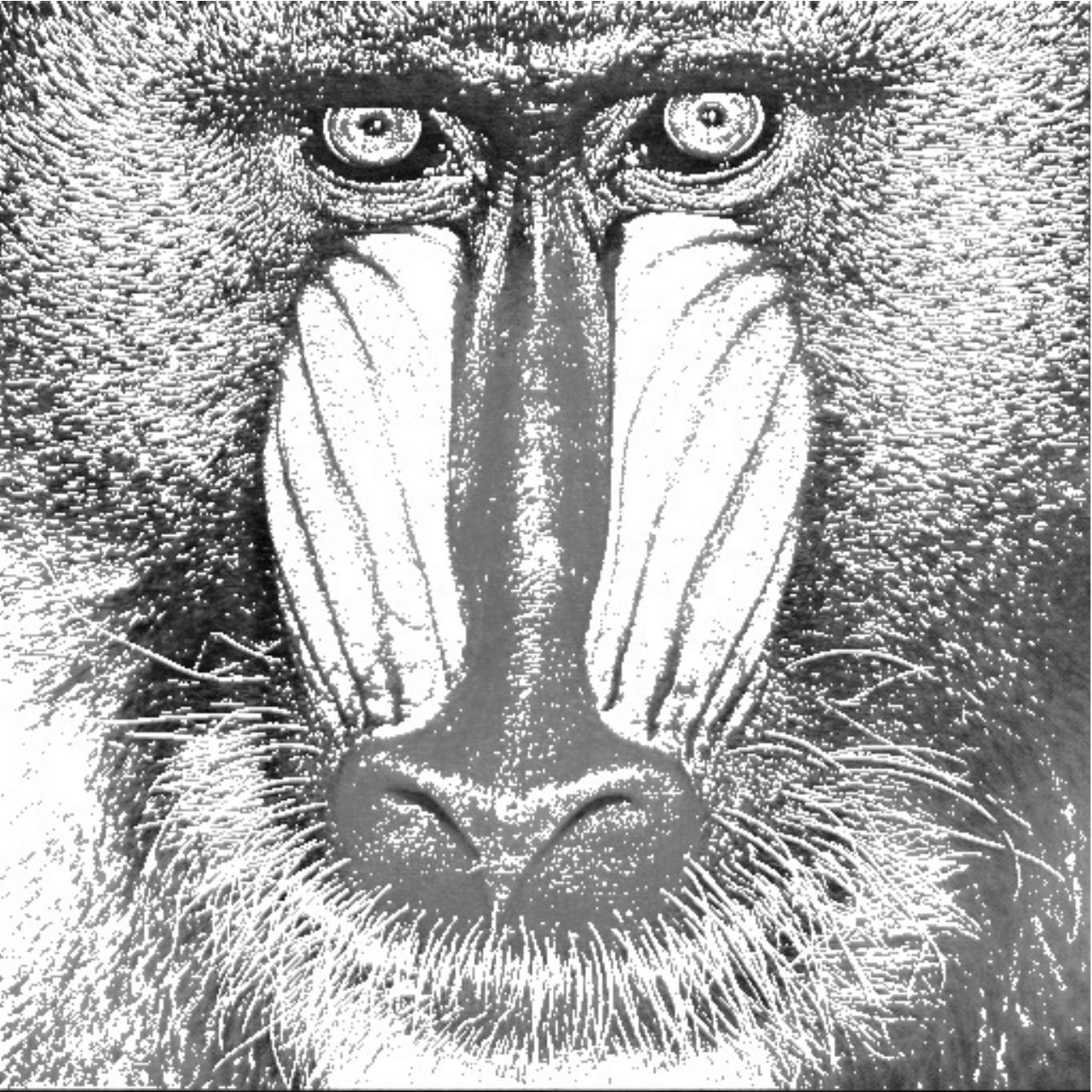}\\
\includegraphics[width=1.7in]{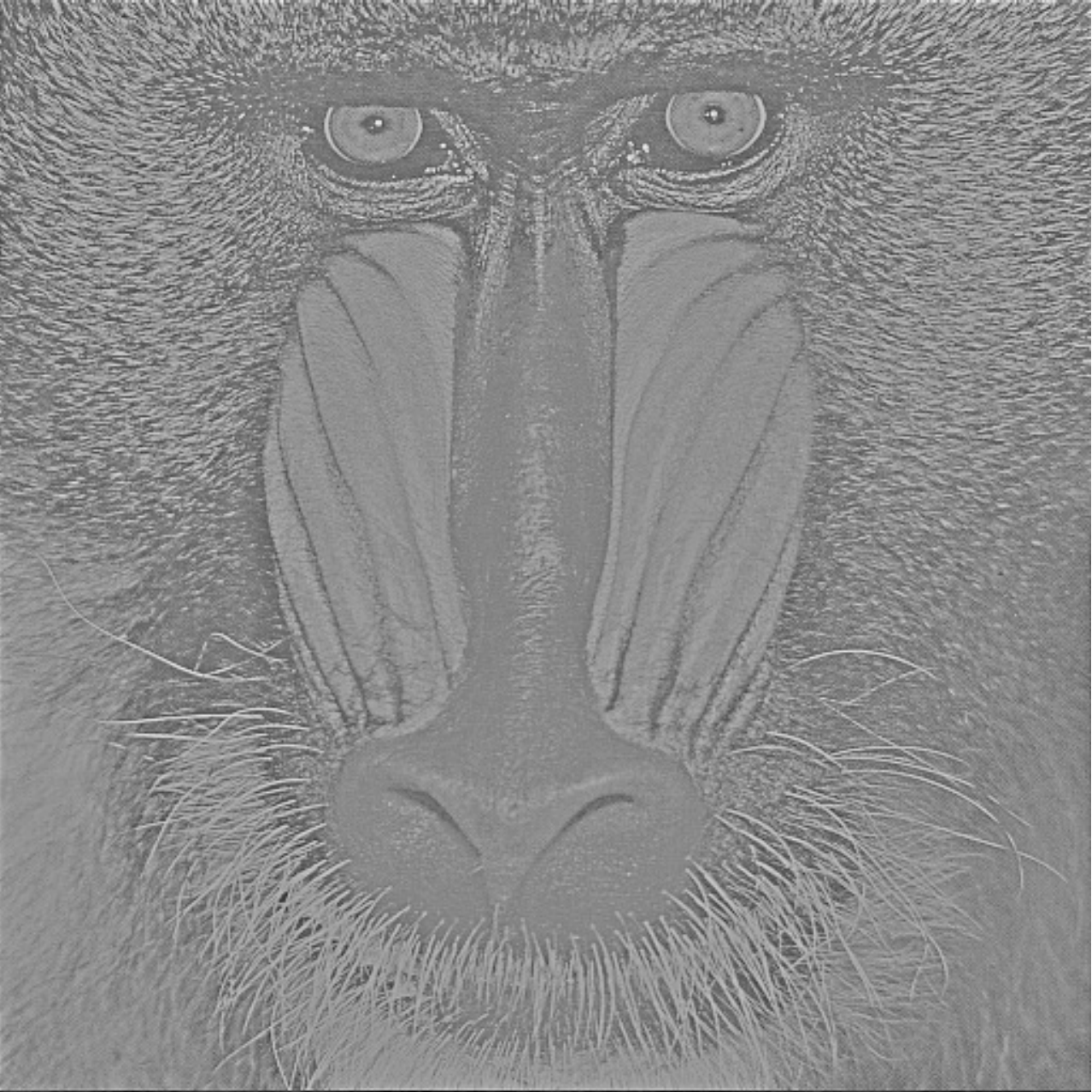}
\includegraphics[width=1.7in]{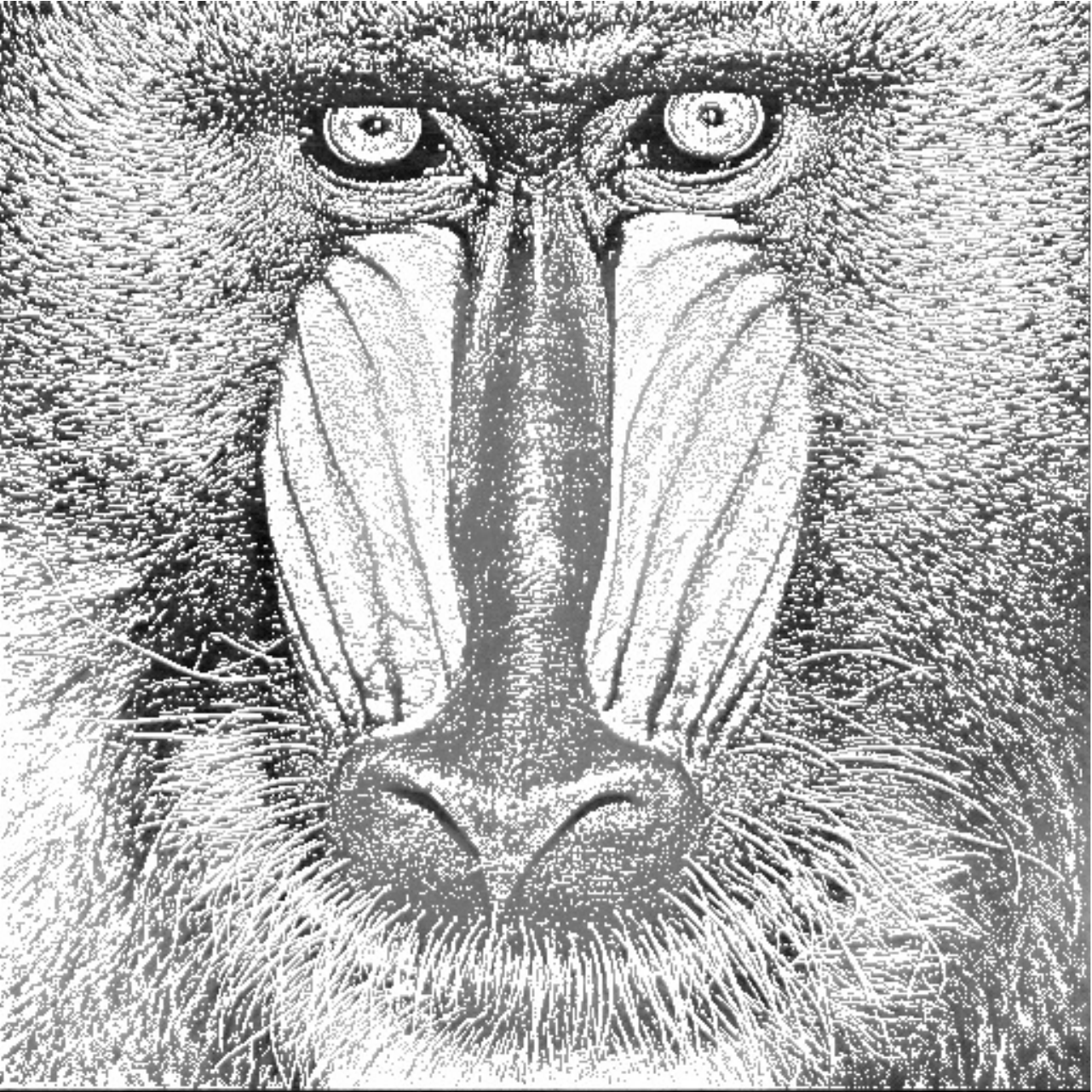} \\
\includegraphics[width=1.7in]{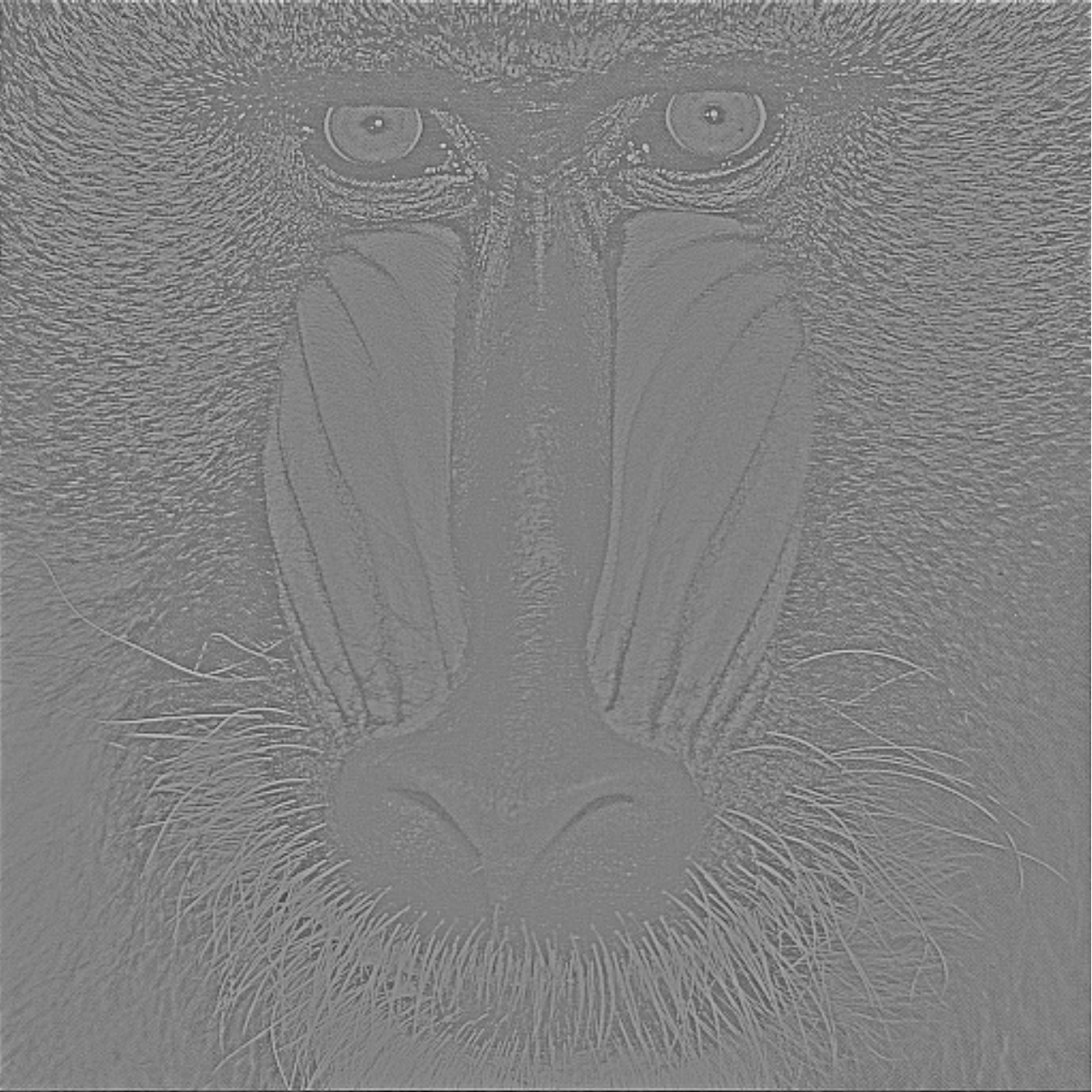}
\includegraphics[width=1.7in]{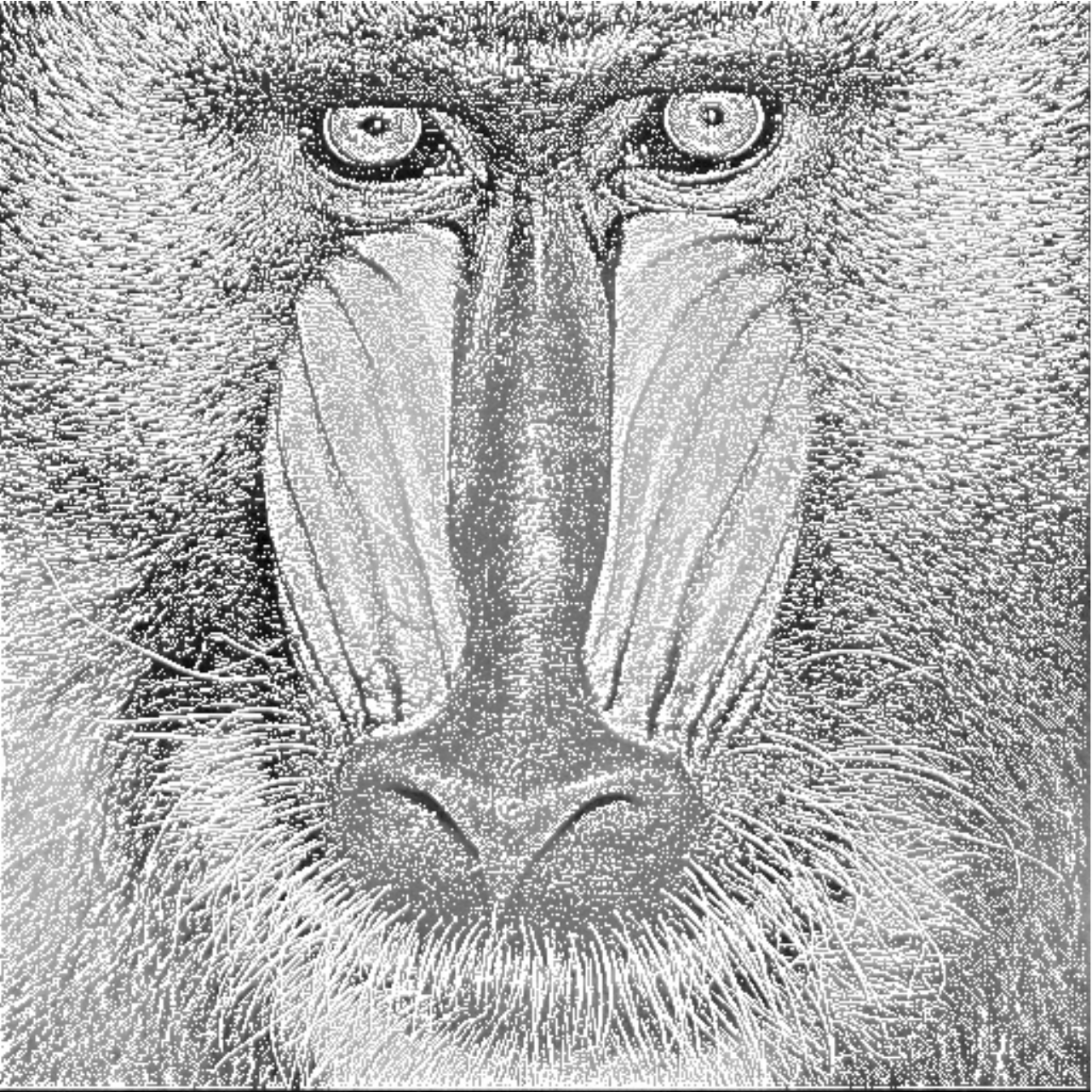}\\
\end{centering}
\caption{Fractional differentiation applied to the Mandrill image
\cite{wolfram} operates analogously to integer-valued derivatives. The left
column shows the fractional derivatives while the right column shows an
edge-detection-like thresholding of the fractional derivative added to the
image (a common way to visualize derivative-like actions on images
\cite{gonzales}). Shown are $q=1$ (the normal derivative, bottom), $q=0.55$
(middle), and $q=0.25$ (top). Different $q$ values show different levels of
detail.}%
\label{fig:mandrilDer}%
\end{figure}

Indeed, frequency domain intuition can show clearly what kinds of signal
processing effects may be expected from fractional-order filters. For example,
applying a fractional integration with order $q \approx1$ to a signal should
be a smoothing operation, a kind of low-pass filter. When applied in two
dimensions to an image, the operation of a $q \approx1$ fractional integration
should be a blurring operation. Similarly, applying a fractional derivative
with order $q \approx1$ to a signal should be a high-pass, high-gain style
operation. When applied in two dimensions to an image, the operation of a $q
\approx1$ fractional derivative should be a sharpening operation such as those
common in edge detectors. Moreover, one should expect that as the order $q$
changes smoothly, the effect of the differintegral operation will change
smoothly. To demonstrate, Figures \ref{fig:mandrilInt} and
\ref{fig:mandrilDer} show various $q$th derivatives and integrals applied to
the same image. The operations of the filters accord reasonably well with the
intuition: the limits as $q$ approaches integer values make sense.

Section \ref{sec:informalDef} presents the key idea of the Riesz
differintegrals as a variation on standard Fourier methods. Section
\ref{sec:compute} shows how the frequency domain definitions can be used to
form the basis of data processing algorithms, and Section
\ref{sec:interpretation} interprets the intuition behind the method. Section
\ref{sec:properties} discusses some interesting and useful properties of the
differintegrals, Section \ref{sec:applications} describes a series of simple
applications, and Section \ref{sec:conclusion} concludes. Computer code to
carry out the required computations are available in both Mathematica and
Matlab at the author's website \cite{differintegralSite}.

\section{Defining Differintegrals via the Fourier Transform}

\label{sec:informalDef}

The Fourier transform of an absolutely integrable function $f(t)$ on the
interval $(-\infty,\infty)$ is a complex-valued function $F (\omega)$ of
frequency $\omega$ defined by
\begin{equation}
\label{eqn:fourier}F (\omega) = \mathcal{F} \{ f(t) \} = \int_{-\infty
}^{\infty} f(t) e^{-j \omega t} dt.
\end{equation}
The inverse Fourier transform can be similarly written as $\mathcal{F}^{-1} \{
F(\omega) \} = f(t) $. A fundamental result relates the time-derivative of the
function to the transform
\begin{equation}
\label{eqn:fourierDif}\mathcal{F} \{ \frac{d f(t)}{dt} \} = j \omega
F(\omega).
\end{equation}
Assuming that all the derivatives, $f^{\prime}(t), \ldots, f^{(n-1)}(t),$
vanish as $t\rightarrow\pm\infty,$ this can be iterated $n$ times to express
the $n$th derivative in terms of the Fourier transform
\begin{equation}
\label{eqn:qthDer}\mathcal{F} \{ \frac{d^{n} f(t)}{dt^{n}} \} = (j \omega)^{n}
F(\omega).
\end{equation}
Similarly, it is possible to express the time-integral of a function in terms
of its Fourier transform. Let $g(t)=\frac{d f(t)}{dt}$, substitute into
(\ref{eqn:fourierDif}), and rearrange to find
\begin{equation}
\label{eqn:fourierInt}\mathcal{F} \{ \int_{-\infty}^{t}g(\tau) d\tau\} =
\frac{1}{j \omega} \mathcal{F} \{ g(t) \} = \frac{1}{j \omega} G(\omega).
\end{equation}
Observe that (\ref{eqn:fourierInt}) is correct only up to a constant since
taking the derivative of $f$ removes any ``DC'' value.

The key idea of the Riesz fractional definition is to rewrite
(\ref{eqn:qthDer}) as
\begin{equation}
\label{eqn:riesz}\mathbf{D}^{q}(f) = \frac{d^{q} f(t)}{dt^{q}} =
\mathcal{F}^{-1} \{ |\omega|^{q} F(\omega) \}
\end{equation}
and to consider this to be a \emph{definition}: the $q$th derivative of the
function $f(t)$ with respect to $t$ is defined to be the inverse Fourier
transform of $|\omega|^{q}$ times the Fourier transform of $f(t)$. The
usefulness of this approach is that $q$ need not be an integer. Moreover, $q$
need not be positive. When $q=-1$, for instance, this recaptures the
relationship in (\ref{eqn:fourierInt}); when $q=1$, (\ref{eqn:riesz})
recaptures (\ref{eqn:qthDer}) (but for the absolute value signs). Hence
(\ref{eqn:riesz}) suffices to define both fractional derivatives (when $q$ is
positive) and fractional integrals (when $q$ is negative).

Some care is needed to make the above argument precise. First, the formal
definition (details can be found in Appendix \ref{sec:appendixDefs}) divides
the Fourier transform into two parts (one from $-\infty$ to $t$ and the other
from $t$ to $\infty$) and it is necessary to replace $(j \omega)$ by
$|\omega|$ to ensure that both integrals converge. Second, the raising of a
number to a fractional power does not have a unique answer, but can result in
multiple possible answers (for example, $n^{1/2}$ can assume two possible
values, one positive and one negative). This can cause sign ambiguities in the
value of the fractional derivatives or integrals. Third, it may be
advantageous in some situations to weight the two halves of the Fourier
transform, as suggested by Feller. In this generalization, (\ref{eqn:riesz})
is replaced by
\begin{equation}
\label{eqn:feller}\mathbf{D}^{q}f = \mathcal{F}^{-1} \{ \left[  c_{1}%
(\theta,q) (j \omega)^{q} +c_{2}(\theta,q) (-j \omega)^{q} \right]  F(\omega)
\}.
\end{equation}
For details, see Appendix \ref{sec:feller}.

\section{Computing Differintegrals}

\label{sec:compute}

\begin{figure}[ptbh]
\begin{center}
\includegraphics[width=3.5in]{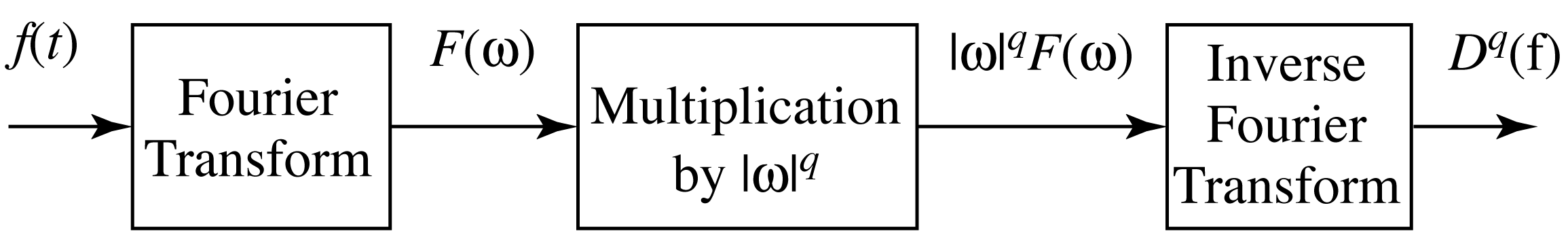}
\end{center}
\caption{The fractional derivative (or integral) of a function $f(t)$ can be
calculated straightforwardly in the frequency domain using
(\ref{eqn:pseudocode}).}%
\label{fig:blockDiagram}%
\end{figure}

The defining equation (\ref{eqn:riesz}) is not only a theoretical definition,
it can also be used as a basis for computation by replacing the Fourier
transform with the Discrete Fourier Transform (DFT). Given a data sequence $f$
of length $n$, let $w$ be a length-$n$ frequency vector spanning normalized
frequency $[-1,1]$. The $q$th fractional differintegral is straightforwardly
implemented in pseudocode as
\begin{equation}
\label{eqn:pseudocode}\mbox{IFFT( Abs(w)}^{\wedge}\mbox{q} \mbox{ FFT(f) )}
\end{equation}
where the power is an element-by-element operation and where FFT and IFFT
represent the DFT and its inverse. This is shown in block diagram form in
Figure \ref{fig:blockDiagram}.

\begin{figure}[ptbh]
\begin{center}
\includegraphics[width=3.5in]{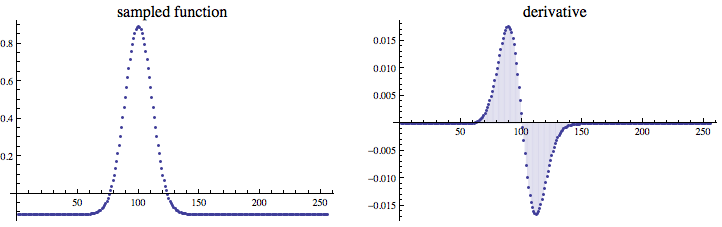}
\includegraphics[width=3.5in]{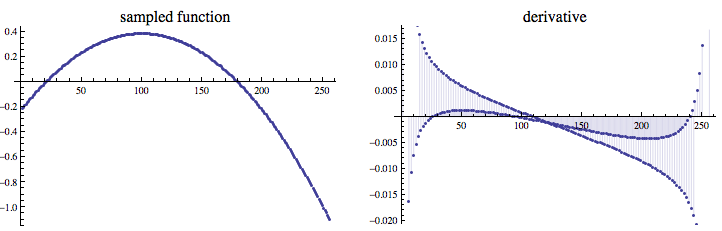}
\end{center}
\caption{Calculating a derivative using the DFT method (\ref{eqn:riesz}%
)-(\ref{eqn:pseudocode}) may give unexpected results (bottom curves) when the
function is not periodic. }%
\label{fig:derivPer}%
\end{figure}

Perhaps the simplest case is $q=1$, using (\ref{eqn:pseudocode}) to calculate
the (regular) derivative. Normally, this would be a waste of computational
effort since there are efficient algorithms for calculating derivatives
(finite differences, differential quadrature, etc.) and (\ref{eqn:pseudocode})
requires two DFTs. But it is worth considering an example that shows how
(\ref{eqn:pseudocode}) assumes periodicity. Figure \ref{fig:derivPer} shows
two examples of derivatives calculated via this method; in the top, the
function is periodic and the derivative appears plausible. For example, it is
positive when the slope of the sampled function is increasing and goes to zero
when the sampled function flattens out. In the bottom example, the derivative
is not (as might be expected) closely related to the derivative of the implied
sampled function (a parabola), but rather is the derivative of one period of
the periodic extension. This same factor occurs in all differintegrals
calculated via the method (\ref{eqn:riesz})-(\ref{eqn:pseudocode}). It can be
ameliorated by windowing (which tapers both ends of the function to zero).

\begin{figure}[ptbh]
\begin{center}
\includegraphics[width=1.6in]{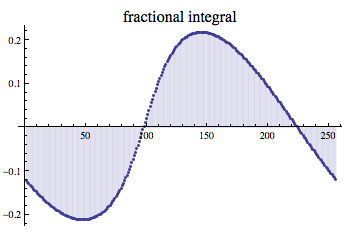}
\includegraphics[width=1.6in]{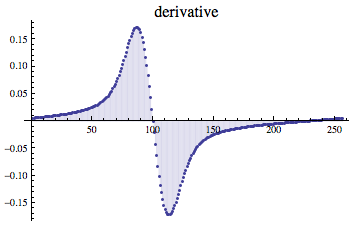} \\[0pt]%
\includegraphics[width=1.6in]{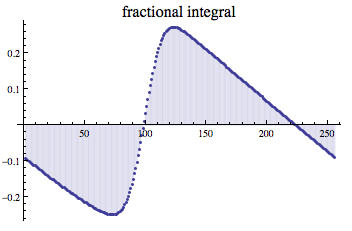}
\includegraphics[width=1.6in]{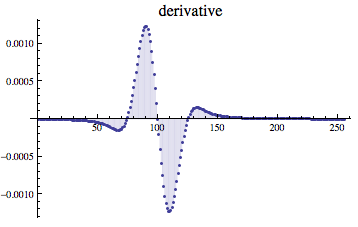} \\[0pt]%
\includegraphics[width=1.6in]{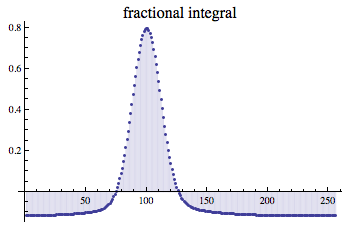}
\includegraphics[width=1.6in]{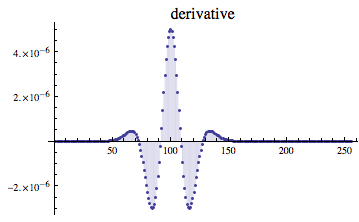}
\end{center}
\caption{Using the sampled function from the top-left of Figure
\ref{fig:derivPer}, several differintegrals are calculated using the DFT
method (\ref{eqn:riesz})-(\ref{eqn:pseudocode}). The left column (from top to
bottom) has integrals with $q=-2.3, -1.05, -0.1$ and skew parameters
$\theta=1,1,0$, while the right column has $q=0.3, 1.8, 3.6 $ and skew
parameters $\theta=1,1,0$. }%
\label{fig:fracderivPer}%
\end{figure}

Figure \ref{fig:fracderivPer} shows various fractional integrals and
fractional derivatives of the sampled function from the top row of Figure
\ref{fig:derivPer}. These also incorporate the Feller ``skew'' parameters
$c_{1}(\theta,q)$ and $c_{2}(\theta,q)$ which weight the contributions from
the two halves of the Fourier transform, replacing (\ref{eqn:riesz}) with
(\ref{eqn:feller}) and (\ref{eqn:pseudocode}) with
\begin{equation}
\label{eqn:pseudocode2}\mbox{IFFT( c1*(j w)}^{\wedge}\mbox{q}
\mbox{+ c2*(-j w)}^{\wedge}\mbox{q} \mbox{ FFT(f) )}
\end{equation}
where c1 and c2 are given by (\ref{eqn:feller2})-(\ref{eqn:fellerCs}) as in
Appendix \ref{sec:feller}.

In one dimension, w (of (\ref{eqn:pseudocode})) is a vector that represents
normalized frequency. In two dimensions, w is a matrix that represents two
dimensions $(\omega_{1}, \omega_{2})$ of normalized frequency, the Abs
function is the norm $\mbox{Abs}(\sqrt{\omega_{1}^{2}+\omega_{2}^{2}})$, and
FFT and IFFT represent the two-dimensional Discrete Fourier transform and its
inverse. Examples in two dimensions are shown in Figures \ref{fig:mandrilInt}
and \ref{fig:mandrilDer}, demonstrating that the methods apply equally well to
images as to one dimensional signals.

Integer-valued derivatives are local, in the sense that the derivative of a
function at a point depends only on the values of the function near that
point. In contrast, differintegrals are nonlocal; the value of the
differintegral at a point depends on all values of the function. This means
that there is no simple time-domain formula (like the discrete difference
operator or the Euler integration formula) that can calculate numerical
differintegrals. Rather, calculation in the time domain requires convolution
with an operator that has the same length as the signal. To calculate the
derivative at every point is thus an $O(n^{2})$ operation. In contrast, the
two FFTs that dominate the calculation of (\ref{eqn:pseudocode}) are $O(n
\log(n))$. Thus the calculation in the Fourier domain is more straightforward
to carry out when compared to the time-domain convolution, and it is also
computationally advantageous.

\section{Interpreting Differintegrals}

\label{sec:interpretation}

\begin{figure}[ptbh]
\begin{center}
\includegraphics[width=1.722in]{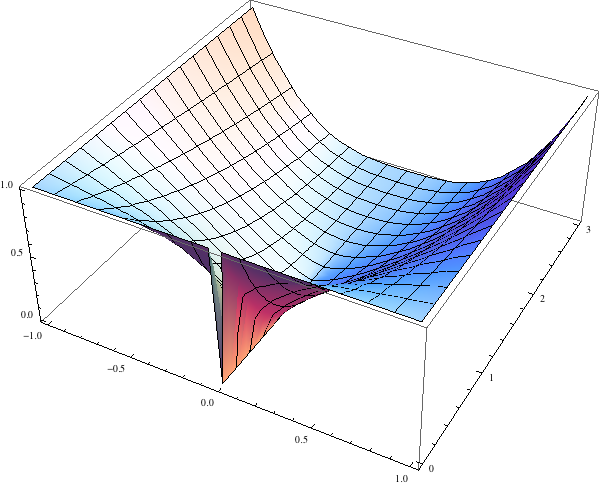}
\includegraphics[width=1.722in]{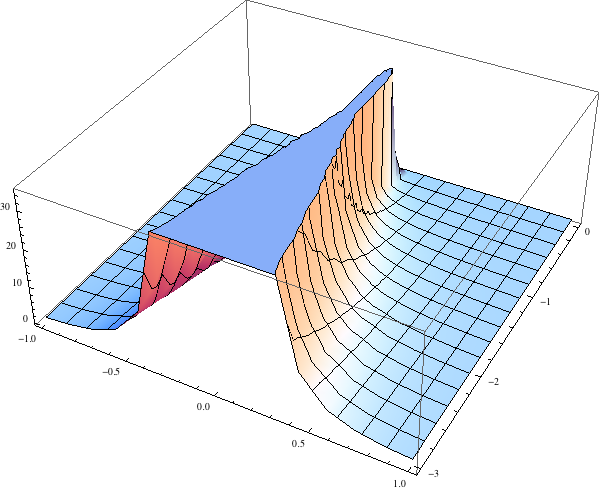}
\end{center}
\caption{Plot of $H(\omega) = |\omega|^{q}$ is divided into two parts. On the
left is $3>q>0$, which shows the frequency scaling accomplished by the
fractional derivatives. On the right, $-3<q<0$, showing the frequency scaling
accomplished by the fractional integrals. These can be interpreted as the
frequency response of the ``system'' defined by the Riesz differintegral
(\ref{eqn:riesz}). }%
\label{fig:freqFun}%
\end{figure}

The defining equation (\ref{eqn:riesz}) can also be used to gain insight into
the meaning of the fractional-order filters defined by applying the
differintegral operations to a signal. The right hand side contains a product
of two frequency domain terms
\[
F(\omega) = \mathcal{F} \{ f(t) \} \mbox{ and } |\omega|^{q}.
\]
The first is the Fourier transform (\ref{eqn:fourier}) of the signal that is
being operated on. The second is frequency, raised to a fractional power.
Figure \ref{fig:freqFun} plots this function for a range of $q$. The plot is
divided into two parts, $q>0$ on the left and $q<0$ on the right. The
frequency axis is normalized to $(-1,1)$.

The convolution property of Fourier transforms relates the convolution
(denoted $\ast$) of two functions to the product of their transforms. Thus
\begin{equation}
\label{eqn:convProperty}\mathcal{F} \{ g(t) \ast h(t) \} = \mathcal{F} \{ g(t)
\}\mathcal{F} \{ h(t)\} = G( \omega) H( \omega).
\end{equation}
In the filtering application, $g(t)$ may be interpreted as a signal and $h(t)
$ may be interpreted as the impulse response of a system. The output of the
system can be calculated either by convolving in the time domain or by
multiplying in the frequency domain. The frequency content of the output is
interpreted as the product of the transform of the input and the frequency
response of the system. In the differintegral setting, $g(t)$ may be
interpreted as the signal to be processed and $h(t)$ is the impulse response
of the system. Since $H(\omega)=|\omega|^{q}$, the impulse response is
$h(t)=\mathcal{F}^{-1} \{ H(\omega) \}=\mathcal{F}^{-1} \{|\omega|^{q} \}$.

For example, consider the fractional order derivative of a function $f(t)$
which has Fourier transform $F(\omega)$. According to (\ref{eqn:riesz}), the
transform of this derivative is the product of $F(\omega)$ and $H(\omega
)=|\omega|^{q}$. A typical contour of such a function is shown in the left
hand side of Figure \ref{fig:freqFun}. It is $1$ at high frequencies and
descends smoothly (at least as long as $q>0$) to zero at $\omega=0$. In words,
this is a kind of highpass filter which passes high frequencies and attenuates
low frequencies. Thus the fractional derivative can be interpreted as highpass operation.

Analogously, consider the fractional order integral of a function $f(t)$ which
has Fourier transform $F(\omega)$. According to (\ref{eqn:riesz}), the
transform of this integral is the product of $F(\omega)$ and $H(\omega
)=|\omega|^{q}$, where now $q<0$. A typical contour of such a function is
shown in the right hand side of Figure \ref{fig:freqFun}. It is $1$ at high
frequencies and ascends continuously towards infinity as $\omega$ approaches
zero. In words, this is a kind of lowpass filter which emphasizes low
frequencies in comparison to high frequencies. Thus the fractional integral
can be interpreted as lowpass operation.

By the convolution property (\ref{eqn:convProperty}), these differintegrals
can be calculated in the frequency domain (as in Section \ref{sec:compute}) or
by calculating $h(t)$ and then convolving in the time domain. Indeed, a
significant amount of effort is required to write $h(t)$ explicitly, and the
formulas in \cite{herrmann} and \cite{mathieu} are complicated, involving
$\Gamma$ functions and collections of factorials. Moreover, the branch cut
problems in these inversions can be formidable, limiting the validity of the
formulas to small ranges of values of $q$.

\section{Properties of Differintegrals}

\label{sec:properties}

Fractional-order derivatives and integrals are closely related to
integer-order derivatives and integrals, in the sense that they share a common
origin (in the Riesz definition (\ref{eqn:riesz}) at least) and a common
interpretation as lowpass and/or highpass filters as given by the frequency
response $H(\omega)$ of Figure \ref{fig:freqFun}. It should come as no
surprise that they also share many properties, and this section details some
of these properties.

\begin{enumerate}
\item \textbf{Linearity:} Differintegrals are linear in the sense that
\begin{equation}
\mathbf{D}^{q}(a_{1}f_{1}+a_{2}f_{2})= a_{1} \mathbf{D}^{q} f_{1} + a_{2}
\mathbf{D}^{q} f_{2},
\end{equation}
where $a_{1}$ and $a_{2}$ are constants and $f_{1}(t)$ and $f_{2}(t)$ are
Fourier integrable functions. This follows immediately from the linearity of
the integrals.

\item \textbf{Composition:} The differintegrals $\mathbf{D}^{q_{1}}$ and
$\mathbf{D}^{q_{2}}$ can be composed so that
\begin{equation}
\label{eqn:composition}\mathbf{D}^{q_{1}}\left(  \mathbf{D}^{q_{2}} f \right)
= \mathbf{D}^{q_{2}} \left(  \mathbf{D}^{q_{1}} f \right)  =\mathbf{D}%
^{q_{1}+q_{2}}f.
\end{equation}
The composition rule can be demonstrated by applying the definition
(\ref{eqn:riesz}) to $\mathbf{D}^{q_{2}}f$, then to $\mathbf{D}^{q_{1}}f$, and
then simplifying
\begin{align}
\mathbf{D}^{q_{1}} \left(  \mathbf{D}^{q_{2}}f \right)   &  =\mathbf{D}%
^{q_{1}} \left(  \mathcal{F}^{-1} \left\{  \left\vert \omega\right\vert
^{q_{2}}F(\omega) \right\}  \right) \nonumber\\
&  =\mathcal{F}^{-1} \left\{  \left\vert \omega\right\vert ^{q_{1}}
\mathcal{F} \left\{  \mathcal{F}^{-1} \{ \left\vert \omega\right\vert ^{q_{2}}
F(\omega) \} \right\}  \right\} \nonumber\\
&  =\mathcal{F}^{-1} \left\{  \left\vert \omega\right\vert ^{q_{1}} \left\vert
\omega\right\vert ^{q_{2}} F(\omega) \right\} \nonumber\\
&  =\mathcal{F}^{-1} \left\{  \left\vert \omega\right\vert ^{q_{1}+q_{2}}
F(\omega) \right\}  =\mathbf{D}^{q_{1}+q_{2}}f.\nonumber
\end{align}
In general, RL differintegrals only commute under special circumstances
\cite{bayin} (having to do with the boundary conditions of the RL Laplace
transform, as discussed in Appendix \ref{app:RieszInt}). In the present case
the required boundary conditions are fulfilled because of the assumption of
the existence of the Fourier transform of the function $f$.

\item \textbf{Identity:} A special case of (\ref{eqn:composition}) is when
$q_{1}=-q_{2}$
\[
\mathbf{D}^{q}\left(  \mathbf{D}^{-q} f \right)  = \mathbf{D}^{-q}\left(
\mathbf{D}^{q} f \right)  = \mathbf{D}^{0} f = f.
\]
For a given $f$, the differintegrals form a commutative group with identity
$\mathbf{D}^{0}$ and where the inverse of element $\mathbf{D}^{q}$ is
$\mathbf{D}^{-q}$. This can be interpreted from a ``block diagram''
perspective as saying that differintegral operators act like linear elements
where differintegral blocks may be combined and rearranged in many of the same
ways that linear time-invariant transfer functions can be combined and rearranged.

\item \textbf{Leibniz's Rule:} The differintegral of the $q$th order of the
multiplication of two functions $f$ and $g$ is given by the formula
\begin{equation}
\mathbf{D}^{q}(f g)=\sum_{k=0}^{\infty} \binom{q}{k} \mathbf{D}^{q-k}(f)
\ \mathbf{D}^{k}(g) ,
\end{equation}
where the binomial coefficients are calculated by replacing the factorials
with the corresponding gamma functions.
\end{enumerate}

In terms of Fourier analysis, differintegrals are a special case of multiplier
operators \cite{duoandikoetxea}, translation-invariant operators that reshape
the frequencies in a function. Many of the computational results of Section
\ref{sec:compute} hold for general multiplier operators and much of the
frequency-domain intuition of Section \ref{sec:interpretation} still apply in
this more general setting, though the details will change to reflect the
specifics of the multiplier under consideration.

\section{Applications}

\label{sec:applications}

This paper began with a series of motivating images processed by the
differintegral operator: Figure (\ref{fig:mandrilInt}) showed the smoothing
operations performed by fractional integrations and Figure
(\ref{fig:mandrilDer}) showed some simple edge detectors based on fractional
derivatives. This section presents a small number of other applications that
might benefit from the use of differintegral filters and illustrates the use
of some of the parameters.

\subsection{Application of Skew Parameter to Embossing}

\begin{figure}[ptbh]
\begin{center}
\includegraphics[width=1.722in]{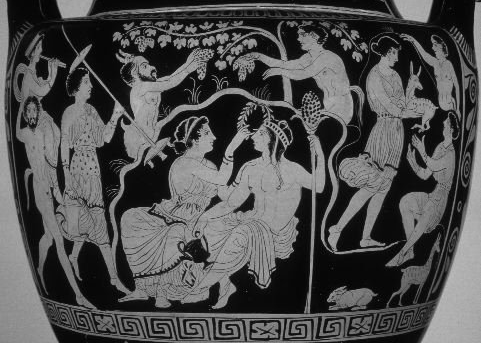}
\includegraphics[width=1.722in]{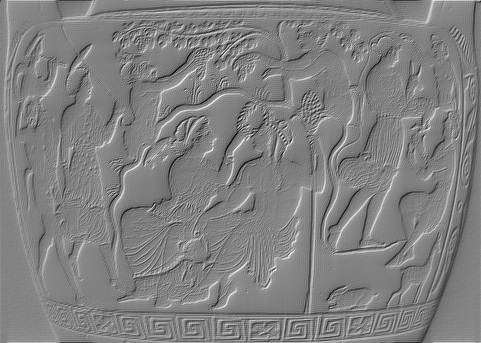}\\[0pt]%
\includegraphics[width=1.722in]{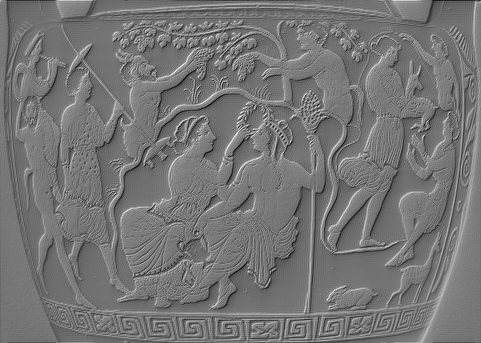}
\includegraphics[width=1.722in]{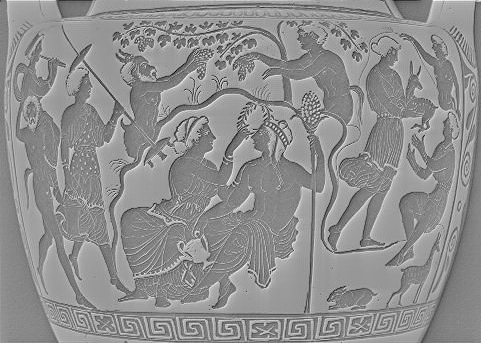}
\end{center}
\caption{The original image of the vase is in the upper left. The fractional
derivative with $q=0.5$ is shown with three different skew values, $\theta=0$
(negative, upper right), $\theta=0.3$ (positive, lower left) and $\theta=0.6$
(negative, lower right). The results are like variations on an embossing
effect.}%
\label{fig:vase}%
\end{figure}

This example fixes the fractional derivative at $q=0.5$ and examines the
effect of several different skew parameters (see (\ref{eqn:feller}) and
(\ref{eqn:fellerCs})) as shown in Figure \ref{fig:vase}. The appearance is
analogous to an embossing effect, which is often accomplished using a
collection of directional derivatives. Here the effect is accomplished by
changing the skew parameter, which weights the contributions of the positive
and negative frequency powers. Changing the skew parameter can make the
embossing effect appear to project either inwards or outwards.

\subsection{Application to Eclipse Detection}

\begin{figure}[ptbh]
\begin{center}
\includegraphics[width=1.7in]{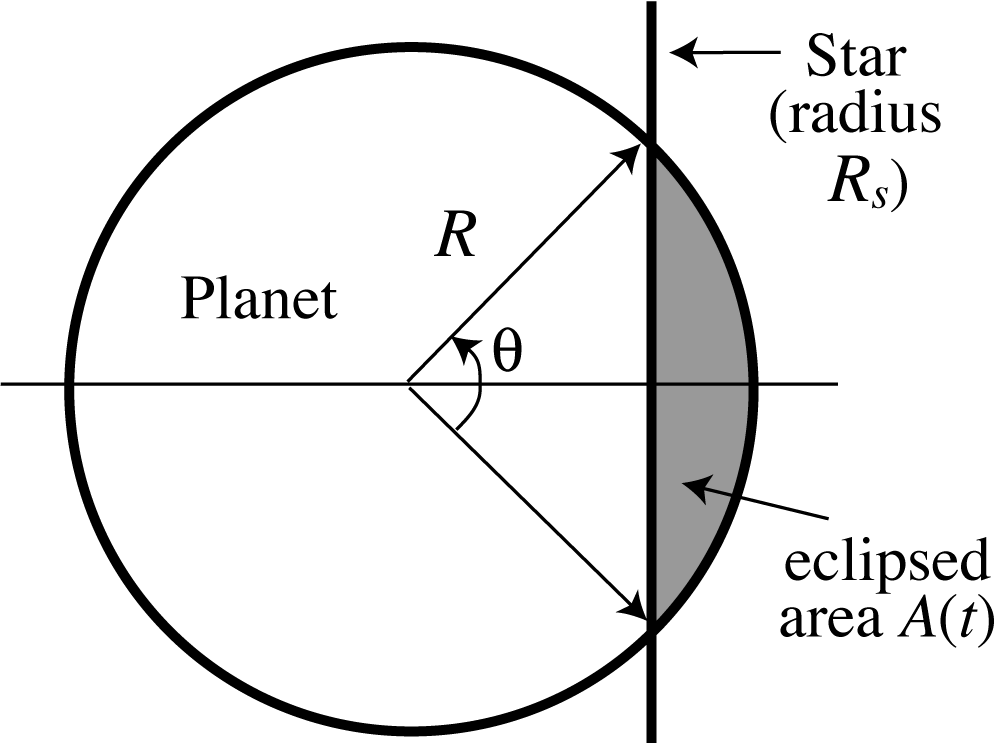}
\end{center}
\caption{The shaded area $A(t)$ of (\ref{eqn:areaEclipse}) is the region
eclipsed by the planet as it traverses the star.}%
\label{fig:eclipse}%
\end{figure}

\begin{figure}[ptbh]
\begin{center}
\includegraphics[width=1.722in]{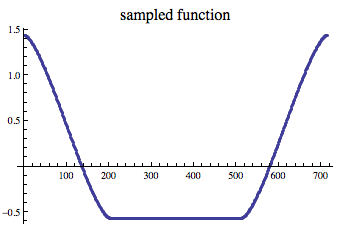}
\includegraphics[width=1.722in]{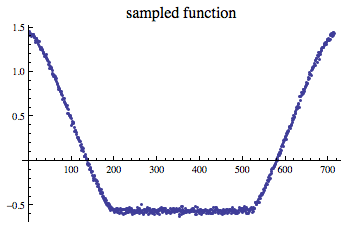}\\[0pt]%
\includegraphics[width=1.722in]{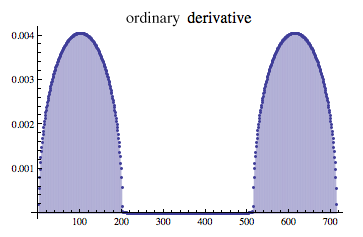}
\includegraphics[width=1.722in]{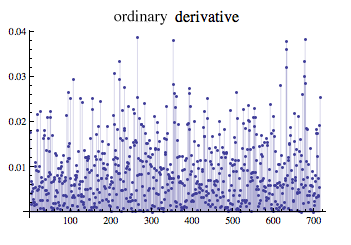}\newline%
\includegraphics[width=1.722in]{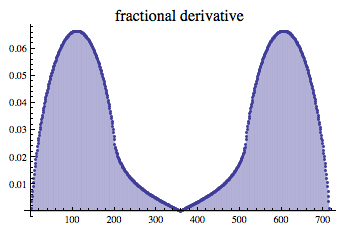}
\includegraphics[width=1.722in]{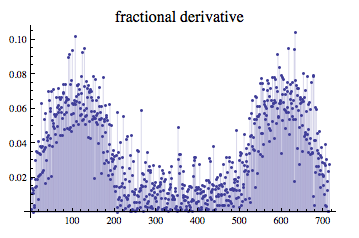}
\end{center}
\caption{The fractional change in brightness (\ref{eqn:area}) is plotted in
the top left for some nominal values of the parameters. A modest amount of
noise (compare to Figure 2 of \cite{kepler}) is added in the top right. The
absolute value of the $q=1$ derivative of the brightness is shown in the
middle left; this has two peaks which show the inflection points of the
brightness curve. The absolute value of the $q=1$ derivative of the noisy
brightness is shown in the middle right; the noise destroys information about
the location of the inflection points. The absolute value of the $q=0.5$
derivative of the brightness is shown in the bottom left; this again has two
peaks which show the inflection points. The absolute value of the $q=0.5$
derivative of the noisy brightness is shown in the bottom right. Even in the
presence of the noise, the approximate location of the inflection points can
still be determined.}%
\label{fig:brightness}%
\end{figure}

An exoplanet transiting a distant star appears as a periodic dip in the
brightness of the star \cite{kepler}. Assuming that the radius $R_{s}$ of the
star is much larger than the radius $R_{p}$ of the planet, the change in the
brightness is approximately proportional to the area
\begin{align}
\label{eqn:areaEclipse}A(t)  &  =\frac{R^{2}}{2}(\theta(t)-\sin\theta(t))\\
\theta(t)  &  = 2\arccos(1-vt/R),\text{ }t\in(0,2r/v)\nonumber
\end{align}
where $R$ is radius of the image of the planet cast on the star, $t$ is the
time, $v$ is the velocity of the planet, and $\theta$ is the angle shown in
Figure \ref{fig:eclipse}. For simplicity, assume that the transit occurs on the equatorial
plane of the star. The radius of the image of the planet on the star is
$R=R_{p}\frac{d_{s}}{d_{p}}$, where $d_{s}$ and $d_{p}$ are the distances of
the star and the planet from the point of observation. After the planet moves
completely within the field of the star, the brightness remains constant at
its minimum level until the planet begins its exit on the other side. The
brightness of a star can be written
\[
b=\frac{\pi R_{s}^{2}I_{s}}{d_{s}^{2}},
\]
where $I_{s}$ is the intensity. The fractional change in brightness is
\begin{align}
\frac{\Delta b}{b}  &  = \frac{\left(  \pi R_{s}^{2}-A(t) \right)  I_{s}
/d_{s}^{2}}{b}\nonumber\\
&  = 1-\frac{A(t)}{\pi R_{s}^{2}}. \label{eqn:area}%
\end{align}
This is plotted in the top left portion of Figure \ref{fig:brightness} and a
modest amount of noise is added in the top right. The first derivative (middle
left) shows the location of the points of inflection because these are the
points where the rate of change is largest. In the noisy version, however, the
locations of the inflection points are overwhelmed by the noisiness of the
derivative. The $q=0.5$ derivative (with a skewness of $1$) is shown in the
bottom two parts. Without noise (bottom left), the location of the inflection
points is clear; with noise (bottom right), the inflection points are still
clear, though the precise locations may be difficult to pinpoint. In this
case, the $q=0.5$ derivative would be preferred to the $q=1$ derivative for
the purpose of locating the points of maximum change. This can be viewed as an
application of the CRONE detector \cite{mathieu} to the brightness function
(\ref{eqn:area}).

\subsection{Processing Colored Images}

One approach to the processing of colored images is to filter the red, green,
and blue (RGB) channels separately. While this can sometimes be effective, the
colors in the output may be different from the colors in the original. When
this is undesirable, a common approach is to translate the RGB channels into
the hue-saturation-brightness (HSB) colorspace, to process the brightness
channel alone, and then convert back to RGB for display. This tends to
preserve the hue and saturation (the ``color'') while changing the brightness.
As with (integer-valued) derivative and integral operators, this can be an
effective means of applying differintegral operations to color images.

A series of such images (with different values of $q$ and different skew
values) are shown at the website \cite{differintegralSite}. These include a
collection of differently smoothed $q<0$ and sharpened $q>0$ versions of the
mandrill image from Figures \ref{fig:mandrilInt} and \ref{fig:mandrilDer}.
Visual effects include smooth blurs, edge-like extractions, and posterizations
(when used in conjunction with binarization) depending on the particular
parameters chosen. Also on the website are several interactive demonstrations,
written in the Wolfram .cdf format (see \cite{differintegralSite} for links).
These can be used inside of Mathematica, or can be used by downloading the
free .cdf player from the Wolfram website. The demonstrations allow the user
to ``play with'' the differintegral operators in a straightforward way.

\section{Conclusions}

\label{sec:conclusion}

Fractional order derivatives and integrals are sensible tools that should be
in the practitioner's signal processing toolbox. While it is unreasonable to
expect miraculous new kinds of processing effects from these tools, they do
provide a logical extension to current techniques. Since derivatives and
integrals are at the heart of many different classical signal processing
algorithms, it is reasonable to ask, in each case, whether the use of
fractional-order filters may enhance these applications. In order to test
whether the methods are useful in a given application, the authors provide
computer code in both Mathematica and in Matlab to easily carry out the
required calculations \cite{differintegralSite}.

\begin{appendix}
\section{Basic Definitions}
\label{sec:appendixDefs}
Appendix \ref{app:RLdef} begins with the Riemann-Liouville (RL) definitions of fractional integrals and derivatives. Transforming into the frequency domain, as in Appendices \ref{app:RieszInt} and
\ref{app:RieszDer}, allows restatement of the definitions that hold under fairly general conditions. This formulation was first described by Riesz \cite{riesz}.
Finally, Appendix \ref{sec:feller} provides a useful extension to the
case where the two parts of the differintegrals are weighted appropriately.
\subsection{RL Definition of Differintegrals} \label{app:RLdef}
The $q$th (right hand) fractional integral of a function is defined to be
\begin{equation} \label{eqn:rightRLInt}
_{a^{+}}\mathbf{D}_{t}^{-q} f(t) =\frac{1}{\Gamma (q)}\int_{a}^{t}(t-\tau
)^{q-1}f(\tau )d\tau
\end{equation}
where $\Gamma (q)$ is the gamma function, $a<t$, and $q>0$.
At first it might seem odd to use the letter $\mathbf{D}^{-q}$ for an integral;
doing so allows a unified notation where a positive exponent means
``derivative'' and a negative exponent means ``integral.''
When $q=1$, this corresponds to the ``regular'' integral from
$a$ to $t$ of the function $f(\tau)$.
The definition of the fractional derivative is less straightforward because
the integral in (\ref{eqn:rightRLInt}) diverges for $q \leq 0$.
This can be addressed as in \cite{bayin} by taking the $n$th (integer)
derivative composed with the $0<n-q<1$ fractional integral
where $n=\lceil q \rceil$ is the smallest integer larger than $q$.
Accordingly, the $q$th (right hand) fractional derivative is defined as
\begin{equation} \label{eqn:rightRLDer}
_{a^{+}}\mathbf{D}_{t}^{q}f(t) =\frac{d^{n}}{dt^{n}} \left(
_{a^{+}} \mathbf{D}_{t}^{q-n} f(t) \right).
\end{equation}
Similarly, The RL (left handed) integral and derivative are
\begin{equation}\label{eqn:leftRLInt}
_{b^{-}}\mathbf{D}_{t}^{-q} f(t)  =\frac{1}{\Gamma (q)}\int_{t}^{b}(\tau
-t)^{q-1}f(\tau )d\tau
\end{equation}%
\begin{equation}\label{eqn:leftRLDer}
_{b^{-}}\mathbf{D}_{t}^{q}f(t) =(-1)^{n}\frac{d^{n}}{dt^{n}} \left(
_{b^{-}}\mathbf{D}_{t}^{q-n} f(t) \right),
\end{equation}
where $t<b$, $n=\lceil q \rceil$, and $q>0$.
The most common values $a=-\infty$ and $b=\infty$ are also called the Weyl
differintegral.
\subsection{Riesz Fractional Integral} \label{app:RieszInt}
The Riesz formula arises from the Fourier transform of the right-hand fractional RL integral
(\ref{eqn:rightRLInt}) with $a=-\infty$
\begin{equation} \label{eqn:infD-q}
_{-\infty }\mathbf{D}_{t}^{-q}g(t)=\frac{1
}{\Gamma (q)}\int_{-\infty }^{t}(t-\tau )^{q-1}g(\tau )d\tau ,\text{ }q>0.
\end{equation}%
To calculate this integral, write the Laplace transform of the function
$h(t)=\frac{t^{q-1}}{\Gamma (q)}$ for $q>0$ as
\[
\pounds \{h(t)\}=\frac{1}{\Gamma (q)}\int_{0}^{\infty
}t^{q-1}e^{-st}dt=s^{-q}.
\]
Next, substitute $s=j \omega $ to obtain the Fourier transform of
\[
h_{+}(t)=\left\{
\begin{tabular}{ll}
$\dfrac{t^{q-1}}{\Gamma (q)},$ & $t>0,$ \\
$0$ & $t\leq 0$%
\end{tabular}%
\ \right.
\]%
which is
\[
\mathcal{F}\{h_{+}(t)\}=(j\omega )^{-q},\text{ }q>0.
\]%
The convolution of $h_{+}(t)$ and $g(t)$ is
\begin{align} \nonumber
h_{+}(t)\ast g(t)& =\int_{-\infty }^{\infty }h_{+}(t-\tau )g(\tau )d\tau \\ \nonumber
& =\frac{1}{\Gamma (q)}\int_{-\infty }^{t}(t-\tau )^{q-1}g(\tau )d\tau \\ \nonumber
& =_{-\infty }\mathbf{D}_{t}^{-q}g(t)
\end{align}%
as in (\ref{eqn:infD-q}). Using the convolution property of the Fourier transform
(\ref{eqn:convProperty}), this becomes
\begin{equation}\label{eqn:+jomegaG}
\mathcal{F}\{_{-\infty }\mathbf{D}_{t}^{-q}g(t)\}=(j\omega )^{-q}G(\omega ),
\mbox{ for } q>0
\end{equation}%
where $G(\omega )$ is the Fourier transform of $g(t)$.
Similarly, the left-hand RL integral (\ref{eqn:leftRLInt}) with $b=\infty$ is
\[
_{\infty }\mathbf{D}_{t}^{-q}g(t)=\frac{1}{\Gamma (q)}\int_{t}^{\infty
}(\tau -t)^{q-1}g(\tau )d\tau ,\text{ }q>0.
\]%
After similar manipulations, this becomes
\begin{equation} \label{eqn:-jomegaG}
\mathcal{F}\{_{-\infty }\mathbf{D}_{t}^{-q}g(t)\}=(-j\omega )^{-q}G(\omega ),%
\text{ }q>0.
\end{equation}%
Summing (\ref{eqn:+jomegaG}) and (\ref{eqn:-jomegaG}) gives
gives
\begin{align} \label{eqn:sumDs} \nonumber
\mathcal{F}\{\left[ _{-\infty }\mathbf{D}_{t}^{-q}
+_{\infty}\mathbf{D}_{t}^{-q} \right] g(t)\}
& =\left[ (j\omega )^{-q}+(-j\omega )^{-q}\right] G(\omega ) \\ \nonumber
& =\left\vert \omega \right\vert ^{-q}\left[ j^{-q}+(-j)^{-q}\right]
G(\omega ) \\
& =\left( 2\cos \frac{q\pi }{2}\right) \left\vert \omega \right\vert
^{-q}G(\omega ).
\end{align}%
The combined expression \cite{elsayed}, which is valid for positive
$q$ with $q\neq 1,3,5,\ldots$ is
\begin{align} \nonumber
\mathbf{D}^{-q}(g)& =\frac{1}{2\Gamma (q)\cos \left( \frac{q\pi }{2}%
\right) }\int_{-\infty }^{\infty }(t-\tau )^{q-1}g(\tau )d\tau  \\ \nonumber
& =\frac{\left[ _{-\infty }\mathbf{D}_{t}^{-q}+_{\infty }\mathbf{D}_{t}^{-q}%
\right] g(t)}{2\cos \left( \frac{q\pi }{2}\right) }\\
& =\mathcal{F}^{-1 }\{ \left\vert \omega \right\vert ^{-q}G(\omega ) \}.
\end{align}%
This is the Riesz fractional integral. The final equality results from
taking the inverse Fourier transform of both sides of (\ref{eqn:sumDs}),
after dividing by the term $2\cos \left( \frac{q\pi }{2}\right)$.
\subsection{Riesz Fractional Derivative} \label{app:RieszDer}
Substituting (\ref{eqn:rightRLInt}) into the RL derivative (\ref{eqn:rightRLDer})
with $a=-\infty $ and $q$ positive gives
\begin{align} \label{eqn:rfd} \nonumber
_{-\infty }\mathbf{D}_{t}^{q}g(t)& =\frac{1}{\Gamma (n-q)}\int_{-\infty
}^{t}(t-\tau )^{-q-1+n}g^{(n)}(\tau )d\tau \\
& =_{-\infty }\mathbf{D}_{t}^{q-n}g^{(n)}(t),\mbox{ for } n-1<q<n,
\end{align}
where $g(t)$ and its derivatives are assumed integrable.
Since $q-n<0,$ (\ref{eqn:+jomegaG}) can be used to write the Fourier transform of
(\ref{eqn:rfd}) as
\begin{align} \label{eqn:frfd} \nonumber
\mathcal{F}\{_{-\infty }\mathbf{D}_{t}^{q}g(t)\}& =(j \omega )^{q-n}\mathcal{F%
}\{g^{(n)}(t)\},\text{ }q>0, \\ \nonumber
& =(j \omega )^{q-n}(j \omega )^{n} G(\omega ) \\
& =(j \omega )^{q} G(\omega ).
\end{align}
Similarly,
\begin{equation} \label{eqn:f2rfd}
\mathcal{F}\{_{\infty }\mathbf{D}_{t}^{q}g(t)\}=(-j \omega )^{q}G(\omega ).
\end{equation}
Combining the results in (\ref{eqn:frfd}) and (\ref{eqn:f2rfd}) gives
\begin{align} \nonumber
\mathcal{F}\{\left[ _{-\infty }\mathbf{D}_{t}^{q}+_{\infty }\mathbf{D}%
_{t}^{q}\right] g(t)\}& =\left[ (j \omega )^{q}+(-j \omega )^{q}\right]
G(\omega ) \\
& =\left( 2\cos \frac{q\pi }{2}\right) \left\vert \omega \right\vert
^{q}G(\omega ).
\end{align}%
For $0<q\leq 2$, $q\neq 1,$ the Riesz fractional derivative is
defined \cite{herrmann} as
\begin{equation} \label{eqn:rieszMin}
\mathbf{D}^{q}(g)=-\frac{\left[ _{-\infty }\mathbf{D}_{t}^{q}+_{\infty }%
\mathbf{D}_{t}^{q}\right] g(t)}{2\cos \left( \frac{q\pi }{2}\right) }.
\end{equation}%
The minus sign in (\ref{eqn:rieszMin}) is
introduced to recover the $q=2$ case
\begin{align} \nonumber
\mathbf{D}^{2}(g)& =-\frac{1}{2\pi }\int_{-\infty }^{\infty }\left\vert
\omega \right\vert ^{2}G(\omega )e^{j \omega t}d\omega \\ \nonumber
& =\frac{1}{2\pi }\int_{-\infty }^{\infty }G(\omega )\left[ \frac{d^{2}}{%
dt^{2}}e^{j \omega t}\right] d\omega \\ \nonumber
& =\frac{d^{2}}{dt^{2}}\left[ \frac{1}{2\pi }\int_{-\infty }^{\infty
}G(\omega )e^{j \omega t}d\omega \right] \\ \nonumber
& =\frac{d^{2}}{dt^{2}}g(t).
\end{align}%
\subsection{The Feller Derivative}
\label{sec:feller}
The linear combination
\begin{equation} \label{eqn:feller2}
\mathbf{D}_{\theta }^{q}f(t)=\left[ c_{1}(\theta ,q)_{-\infty }\mathbf{D}%
_{t}^{q}+c_{2}(\theta ,q)_{\infty }\mathbf{D}_{t}^{q}\right] f(t)
\end{equation}%
where
\begin{eqnarray} \nonumber
c_{1}(\theta ,q) &=&-\frac{\sin ((q+\theta )\pi /2)}{\sin \pi \theta }\\ \label{eqn:fellerCs}
c_{2}(\theta ,q) &=&-\frac{\sin ((q-\theta )\pi /2)}{\sin \pi \theta }
\end{eqnarray}%
has been introduced by Feller as a generalization of fractional
derivatives. This weights the left and right-hand differintegrals according
the $c$ parameters and allows extra flexibility in the calculations.
The parameter $\theta$ is called the phase or the {\em skew} factor.
Two special cases are of note:
\begin{enumerate}
\item For $\theta =0$, $c_{1}(0,q)=c_{2}(0,q)=-\frac{1}{2\cos (q\pi /2)}$,
and the Feller derivative $\mathbf{D}_{\theta=0}^{q}(f)$ reduces to the Riesz derivative
$\mathbf{D}^{q}(f)$.
\item For $\theta =1$,
$c_{1}(1,q)=-c_{2}(1,q)=-\frac{1}{2\sin (q\pi /2)}$, and
the Feller derivative combines the left and right-handed derivatives
with opposite signs
\[
\mathbf{D}_{\theta=1}^{q}(f) =\frac{\left[ _{-\infty }\mathbf{D}%
_{t}^{q}-_{\infty }\mathbf{D}_{t}^{q}\right] }{2\sin (q\pi /2)}f(t).
\]
Up to a constant, this case is the same as the ``CRONE detector''
derived in \cite{mathieu} in the time domain.
\end{enumerate}
\begin{figure}[htbp]
\begin{center}
\includegraphics[width=1.722in]{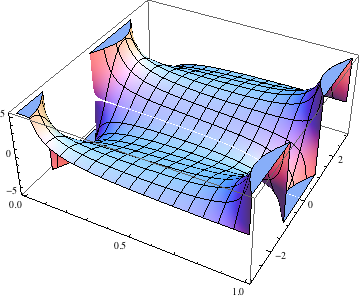}
\includegraphics[width=1.722in]{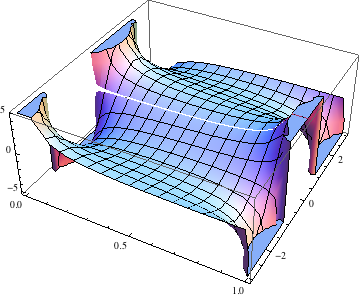}
\end{center}
\caption{Plot of the weighting coefficients
$c_{1}(\theta,q)$ and $c_{2}(\theta,q)$ for $-3<q<3$ and $0<\theta<1$.}
\label{fig:cMcP}
\end{figure}
\end{appendix}

\end{document}